%% file: main.tex
\documentclass[10pt,twocolumn,letterpaper]{article}

\usepackage{iccv}
\usepackage{times}
\usepackage{epsfig}
\usepackage{graphicx}
\usepackage{amsmath}
\usepackage{amssymb}
\usepackage{booktabs}
\usepackage{mathtools}
\usepackage{multirow}
\usepackage{makecell}

\usepackage{colortbl}
\usepackage{bbding}
\usepackage{pifont}
\usepackage{enumitem}
\definecolor{Gray}{gray}{0.9}
\newcommand \footnoteONLYtext[1]
{
	\let \mybackup \thefootnote
	\let \thefootnote \relax
	\footnotetext{#1}
	\let \thefootnote \mybackup
	\let \mybackup \imareallyundefinedcommand
}

\usepackage[pagebackref=true,breaklinks=true,letterpaper=true,colorlinks,bookmarks=false]{hyperref}

\iccvfinalcopy 

\def\iccvPaperID{5620} 

\ificcvfinal\fi

\begin{document}


\title{
No-frills Temporal Video Grounding: \\Multi-Scale Neighboring Attention and Zoom-in Boundary Detection
}
\author{Qi Zhang, Sipeng Zheng , Qin Jin\\
School of Information, Renmin University of China\\
\texttt{\{zhangqi1996, zhengsipeng, qjin\}@ruc.edu.cn}}

\maketitle
\ificcvfinal\thispagestyle{empty}\fi

\input{section/0-abstract}
\input{section/1-introduction}
\input{section/2-related}
\input{section/3-revisit}
\input{section/4-method}

\input{section/5-experiment}
\input{section/6-conclusion}

{\small
\bibliographystyle{ieee_fullname}
\bibliography{egbib}
}

\end{document}

%% file: section/0-abstract.tex
\begin{abstract}
Temporal video grounding (TVG) aims to retrieve the time interval of a language query from an untrimmed video. A significant challenge in TVG is the low "Semantic Noise Ratio (SNR)", which results in worse performance with lower SNR. Prior works have addressed this challenge using sophisticated techniques. In this paper, we propose a no-frills TVG model that consists of two core modules, namely multi-scale neighboring attention and zoom-in boundary detection. The multi-scale neighboring attention restricts each video token to only aggregate visual contexts from its neighbor, enabling the  extraction of the most distinguishing information with multi-scale feature hierarchies from high-ratio noises. The zoom-in boundary detection then focuses on local-wise discrimination of the selected top candidates for fine-grained grounding adjustment. With an end-to-end training strategy, our model achieves competitive performance on different TVG benchmarks, while also having the advantage of faster inference speed and lighter model parameters, thanks to its lightweight architecture.

\end{abstract}
\footnoteONLYtext{The initial NA version  was among the final winners for Natural Language Queries Challenge in IEEE / CVF Computer Vision and Pattern Recognition Conference Ego4D Workshop, 2022.}

%% file: section/1-introduction.tex
\section{Introduction}
\begin{figure}[t]
\begin{center}
	\includegraphics[width=1\linewidth]{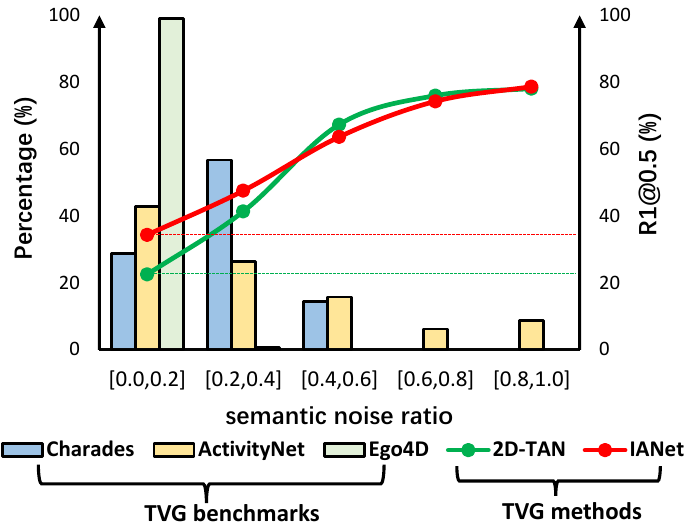}
\end{center}
\caption{
\textbf{Bar}: Statistics on the distribution of SNR across different TVG benchmarks. \textbf{Curve}: Rank1@0.5 accuracy of 2D-TAN and IANet under different SNR on ActivityNet. 
}
\label{fig:intro1}
\end{figure}

Temporal video grounding (TVG) \cite{gao2017CTRL,anne2017TVG1,zheng2022exploring} involves retrieving a corresponding video clip (e.g., represented by a frame interval $[t_s, t_e]$) from an untrimmed video (e.g., represented by a frame interval $[0, T]$) in response to a language query.
Typically, the semantics of the query are related to only a small portion of the long video. 
As such, the TVG task is characterized by a low semantic noise ratio (SNR), which is defined as the ratio of the target clip interval to the untrimmed video interval, i.e. $\mathrm{SNR}=\frac{(t_e-t_s)}{T}$.
As shown in Figure~\ref{fig:intro1}, a significant proportion of TVG trials across different benchmarks suffer from low SNR. 
For example, in Ego4D-NLQ \cite{grauman2022ego4d}, almost all trials have an SNR lower than 0.2. 
Such low SNR property is at the heart of the challenges in the TVG task, as it is extremely difficult for a TVG model to strike a balance between capturing detailed spatial-temporal visual information and suppressing noise under such conditions.
As shown in Figure~\ref{fig:intro1}, when SNR drops from [0.4,0.6] to [0.2,0.4], it leads to a nearly $30\%$ performance drop on the R1@0.5 metric for state-of-the-art methods \cite{liu2021IANET,zhang20192DTAN}.

\begin{figure}[t]
	\begin{center}
		\includegraphics[width=0.9\linewidth]{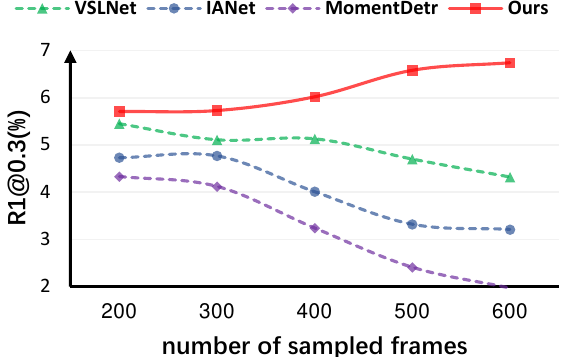}
	\end{center}
	\caption{Rank1@0.3 accuracy results using different number of sampled frames on Ego4D-NLQ. }
	\label{fig:ego4d}
\end{figure}

In contrast to trimmed video tasks such as text-video retrieval~\cite{lin2022swinbert}, which generally have high SNR, tasks with low SNR such as TVG require a more nuanced approach. 
Dense frame sampling can effectively enhance video understanding in high SNR scenarios, however, increasing frame sampling in low SNR tasks may actually have a negative impact on the performance. 
This is due to the semantic asymmetry between the query and the untrimmed video, which is further exacerbated by the low SNR.
As shown in Figure~\ref{fig:ego4d}, current TVG methods do not experience an improvement in performance when the frame sampling is increased; in fact, the performance usually decreases. 
The best results are actually achieved through sparse sampling (200 frames, 0.4 FPS). 
This is paradoxical, as sparse sampling inevitably leads to the loss of important information.

Recently, various sophisticated techniques have been proposed to mitigate the negative effects of low SNR in TVG tasks. 
For instance, MSA~\cite{zhang2021MSA} introduces a multi-stage aggregated transformer to enhance the iterations and alignments between visual and text elements. 
In addition, IANet~\cite{liu2021IANET} adopts a multi-step calibration module with extra learnable parameters to address the nowhere-to-attend problem of non-matched frame-word pairs. MMN~\cite{wang2022MMN} designs metric-learning framework to exploit positive and negative matching samples between video and query. SSRN~\cite{zhu2023SSRN} introduces siamese sampling mechanism to address the limitation of the boundary-bias.
However, all of these techniques require intricate architectures or additional parameters that must be meticulously tuned, which hinders their generalization ability.

In this study, we propose a no-frills model for temporal video grounding that avoids the need for complex techniques. 
Our approach is inspired by human cognitive habits for grounding a video clip. 
Specifically, we humans tend to first skim the video for the target clip based on the local view and then scrutinize the clip boundary carefully.
To achieve this, we propose two core modules in our no-frills TVG model, namely multi-scale neighboring attention (MNA) and zoom-in boundary detection (ZBD), which are modest and easy to plug into other existing models. With only light-weight transformer architecture and multi-layer perceptrons (MLPs), our model still achieves competitive performance on a broad range of TVG benchmarks with 2X inference speedup and 75\% parameter savings.

To be specific, the proposed multi-scale neighboring attention restricts each video token to only aggregate visual contexts from its neighbors, while hierarchically reducing its scope from bottom to top.
It creates a multi-scale pyramid of feature aggregation inside the transformer block, enabling our model to extract the most distinguishing information with multi-scale feature hierarchies from high-ratio noises while refining the extracted semantics throughout the input stream. 
Moreover, our proposed zoom-in boundary detection module screens out the top candidates and then zooms in to focus on local-wise discrimination for fine-grained grounding adjustment. 
This leads to an improvement in model performance with increasing frame sampling rate, as shown in Figure~\ref{fig:ego4d}, distinguishing our method from other approaches.

The main contributions of this work include: 
1) We propose a simple but effective framework for temporal video grounding without bells and whistles.
2) We propose multi-scale neighboring attention and zoom-in boundary detection to reduce the detrimental effect of noises caused by low SNR in untrimmed videos, thereby producing high-quality proposals.
3) We achieve competitive performance on three widely used benchmarks, ActivityNet, Charades and Ego4D-NLQ, while gaining advantages in inference speed and model parameters.

%% file: section/2-related.tex
\section{Related Works}
\input{imgs/overview} 
Early works~\cite{gao2017CTRL,anne2017TVG1} for temporal video grounding (TVG) generally adopt a sliding-window approach. 
These methods involve employing temporal windows to generate a large number of overlapping boundary proposals, followed by a matching module to compute the similarity between the query and proposals. 
However, such a sliding-window mechanism leads to high computational costs. 
Therefore, non-sliding-window models are proposed~\cite{zhang20192DTAN,zhang2020VLSNET,liu2021IANET,liu2020CSMGAN,liu2021CBLN,lei2021MomentDTER}, which first take both video and query as input, then adopt a cross-modal alignment module to jointly fuse visual-textual representations, and finally predict proposals all at once via a detection head. 

Most previous TVG works focus on improving joint visual-textual alignment.
Instead of adopting simple operation such as Hadamard product or concatenation \cite{zhang20192DTAN,wang2022MMN, zhang2021ms2dtan}, these works \cite{li2022VISA, liu2021IANET, lei2021MomentDTER} propose a multi-step iterative structure to better understand the cause and effect of human activities. 
For example, Li \etal ~\cite{li2022VISA} leverage a graph neural network~\cite{scarselli2008GNN} to learn the cross-graph relationships between modalities. 
Liu \etal \cite{liu2021IANET} propose iterative 
inter-modal and intra-modal fusion modules to progressively align complicated semantics between visual and language cues. 
Inspired by the great success of vision transformer \cite{vaswani:transformer} in trimmed video tasks like video captioning \cite{zhang2022DVC, li2020hero, xu2021vlm} and video retrieval \cite{chen2020fine,dzabraev2021mdmmt,gabeur2020multi,liu2019use,liu2021hit,luo2020univl,liu2022ts2net}, Lei \etal \cite{lei2021MomentDTER} introduce a transformer-based TVG framework. 

In addition to improving cross-model alignment, some works have taken another line of research  and focused on upgrading the boundary detection manner.
Zhang \etal ~\cite{zhang20192DTAN} design a novel anchor generation mechanism that adopts a 2D map to model temporal relations.
One dimension of the 2D map represents the starting timestamp while the other represents the end.  
Meanwhile, Lei \etal~\cite{lei2021MomentDTER} propose an anchor-free method based on DETR~\cite{DETR, DeDETR} using query embeddings to predict temporal boundaries directly.

Although these methods have shown promising improvement on TVG benchmarks, they also bring increasingly more modules or techniques to the vanilla framework, making it more sophisticated and difficult to reproduce.
Therefore, the goal of our study is to propose a no-frill framework that offers \textit{competitive performance}, \textit{better efficiency}, and \textit{plug-able modules} without bells and whistles.

%% file: imgs/overview.tex
\begin{figure*}[ht]
	\begin{center}
		\includegraphics[width=1.\linewidth]{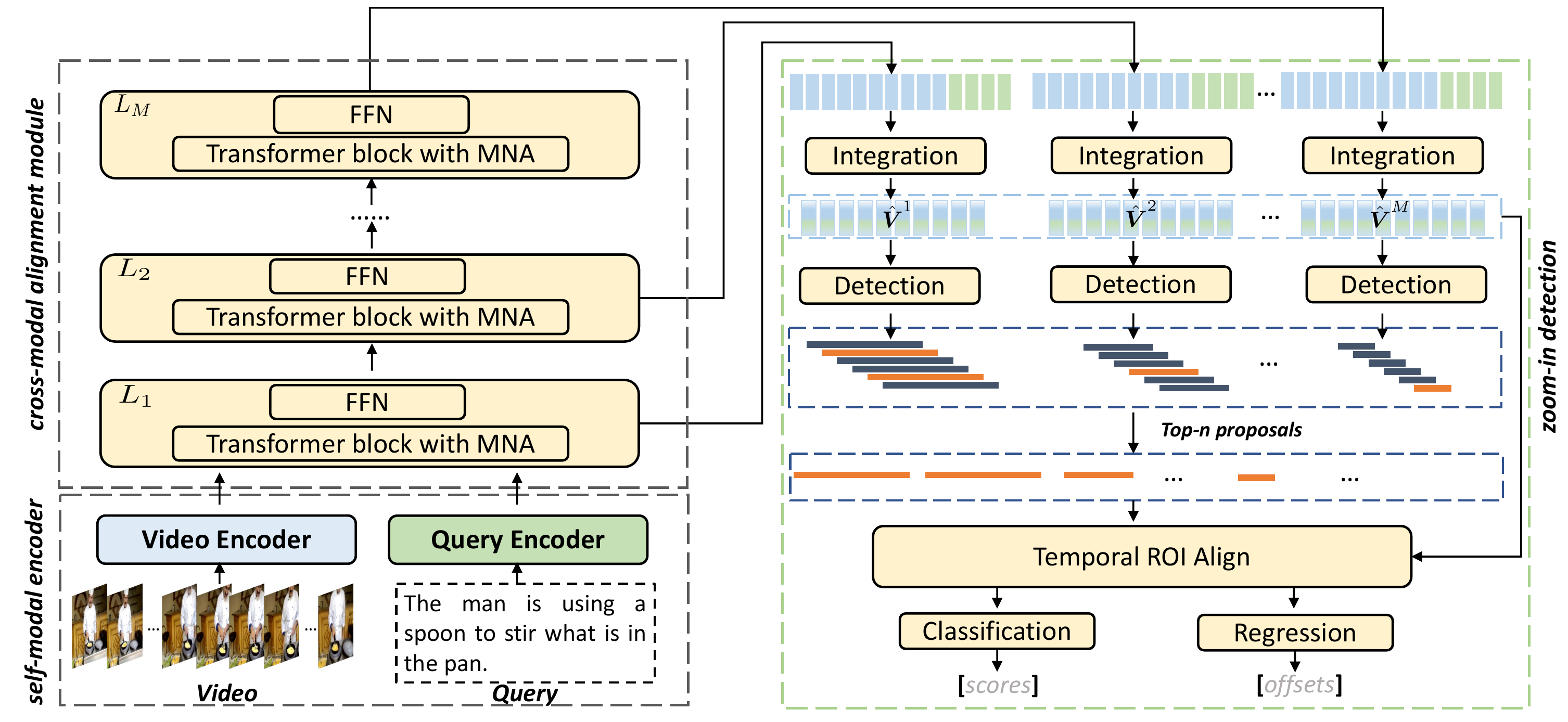}
	\end{center}
	\caption{Our framework comprises three key components: 
	(1) self-modal encoder; 
	(2) cross-modal alignment module enhanced by multi-scale neighboring attention (MNA);
	(3) zoom-in boundary detection (ZBD) that utilizes  multi-scale hierarchy.}
	\label{fig:model}
 \vspace{-5pt}
\end{figure*}

%% file: section/3-revisit.tex
\section{Revisiting Previous TVG Framework}
\input{imgs/local_attn}
Most TVG methods
\cite{zhang2020VLSNET,liu2021IANET,liu2020CSMGAN} consist of three key components: (1) single-modal video and text encoder for intra-modality learning; (2) cross-modal alignment module to jointly learn visual and textual semantics; (3) boundary detection head to provide proposals of the given query.

\noindent\textbf{Video and Text Encoder. }
Given a video, the model first samples $T$ frames by a certain ratio and extracts a sequence of frame representations using off-the-shelf backbones such as C3D~\cite{Tran_2015_ICCV_C3D} or I3D~\cite{Carreira_2017_CVPR_I3D}.
These representations are added by positional embeddings to retain their temporal order information, which are then fed into the video-encoder to generate $D$-dimensional video embeddings $\boldsymbol{V}=\{v_i\}_{i=1}^{T}\in \mathbb{R}^{T \times D}$.
Meanwhile, the language query is tokenized into $L$ tokens, and their corresponding token embeddings are obtained using a pre-trained linguistic model such as Glove \cite{pennington2014GLOVE}. 
Then a text encoder like bidirectional GRU~\cite{cho2014GRU} is adopted to encode the text embeddings as $\boldsymbol{Q}=\{q_i\}_{i=1}^{L}\in \mathbb{R}^{L \times D}$.

\noindent\textbf{Cross-modal Alignment Module. }
After encoding the video and text modality, the model concatenates $\boldsymbol{V}\in \mathbb{R}^{T \times D}$  and $\boldsymbol{Q}\in \mathbb{R}^{L \times D}$ into $\boldsymbol{X}=\{x_i\}_{i=1}^{T+L}\in \mathbb{R}^{(T+L) \times D}$, which serves as the input to the cross-modal alignment module. 
As illustrated in Figure~\ref{fig:local_attention} (a), most existing methods adopt a transformer architecture to
aggregate inter-modal information from the global visual-text context with full attention. 
Given the input feature map $\boldsymbol{X}\in \mathbb{R}^{(T+L) \times D}$, let $s_i$ denotes the $i$-th context feature of a query element, the full-attention feature aggregation is computed by:

\begin{equation}
\text{FullAttn}(s_i, \boldsymbol{X}) = \sum_{k=1}^{T+L} A_{ik} \mathbf{W}x_k,
\label{eq:full_attn}
\end{equation}
where $A_{ik}$ refers to the attention weight of $k$-th key for $i$-th query element. 
$\mathbf{W}$ is the projection matrix for key elements.

\noindent\textbf{Boundary Detection Head. }
The anchor-based detection we adopted in this paper is one typical boundary detection approach.
For a given video with $T$ sampled frames, the model pre-defines $H$ anchors on each frame with distinct scales which results in $HT$ anchors in total. 
The timestamp of the $h$-th anchor on the $t$-th frame can be denoted as $[t-w_h, t+w_h]$, where $w_h$ is the anchor radius. 
Then, the boundary detection head employs two MLPs to predict anchor coordinate regression $o \in [1, T]^{2HT}$ and confidence score $\tau \in [0,1]^{HT}$.  

%% file: imgs/local_attn.tex
\begin{figure*}[ht]
	\begin{center}
		\includegraphics[width=0.95\linewidth]{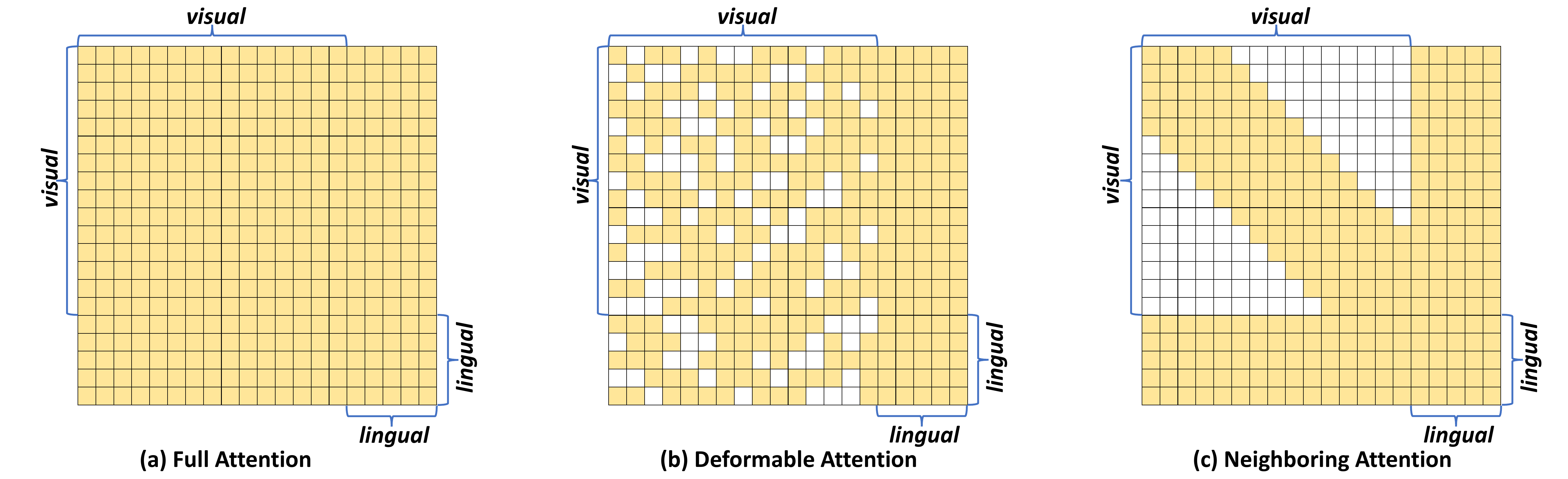}
	\end{center}
	\caption{Different types of attention mechanism in the transformer block, including our newly proposed multi-scale neighboring attention (MNA). The \textbf{visible key} elements in attention calculation are highlighted in \textbf{yellow}. The vertical coordinate represents the query elements and
	the horizontal coordinate represents the key.  }
	\label{fig:local_attention}
 \vspace{-8pt}
\end{figure*}

%% file: section/4-method.tex
\section{No-frills Temporal Video Grounding}

Due to the interference of low SNR, current methods struggle with aligning the semantics of the query with the related clip in the untrimmed video. 
To tackle this issue, we propose two modules: multi-scale neighboring attention (MNA) and zoom-in boundary detection (ZBD). 
The main idea of our method is to filter out discriminative information related to the query from distracting visual noises, thereby reducing the impact of low SNR.
Figure~\ref{fig:model} illustrates the overall structure of our framework.

\subsection{Multi-scale Neighboring Attention (MNA)}

The traditional full-attention operator falls short for TVG due to the excessive visual redundancy found in untrimmed videos.
In response, we first introduce Multi-scale Neighboring Attention (MNA), a self-attention operator that enhances the aggregation of key elements while decreasing visual redundancy in a transformer block.
In contrast to commonly used local-wise operators such as deformable attention \cite{DeDETR} in Figure~\ref{fig:local_attention} (b), which aggregates features by sampling a fraction of key points around a reference from the entire feature set, our MNA extracts contexts of key elements by considering their continuous neighboring frames.
The concept behind such a local-wise operator is based on the nearest neighboring property, which assumes that one visual element normally has the strongest semantic bond to its closest neighbors.
However, when SNR is low, it becomes difficult to select query-relevant contexts from the entire long video.
To address this, our MNA begins aggregation from local neighbors and then zooms in the scope through a multi-scale pyramid to gradually refine the semantics from the network's input to its output.

\noindent\textbf{Neighboring Attention Operator (NA). } 
For each visual query $s_i^v$ in the feature map $\boldsymbol{X}$, we adopt local masking to restrict only its neighboring visual tokens to be visible.
As shown in Figure~\ref{fig:local_attention} (c), we first compute the dot products between the query of $i$-th visual token and a set of key elements.
The key element set can be represented as: 

\begin{equation}
\Omega = \{x_k\}_{k=(i-r)}^{i+r}\cup \{x_k\}_{k=T}^{T+L}
\label{eq:ne_attn_range}
\end{equation}
where $r$ denotes the temporal window radius.
Given a text token $s_i^l$, we adopt the full-attention operator for its query. 
Then, our neighboring attention (NA) is computed as:

\begin{equation}
\text{NA}(s_i^v, \boldsymbol{X}) = \sum_{k\in\Omega} A_{ik} \mathbf{W}x_k,
\label{eq:ne_attn1}
\end{equation}

\begin{equation}
\text{NA}(s_i^l, \boldsymbol{X}) = \sum_{k=1}^{T+L} A_{ik} \mathbf{W}x_k,
\label{eq:ne_attn}
\end{equation}

Our proposed neighboring attention (NA) operator will be directly plugged into each transformer block in the cross-modal alignment module.

\noindent\textbf{Multi-scale Hierarchies.} 
Moreover, to enable our model with flexible scope modeling, we upgrade the vanilla NA operator with multi-scale hierarchies.
The main idea is to progressively reduce the temporal scope from bottom to top while maintaining a consistent temporal resolution (i.e. sequence length) across the network.
To be specific, each layer's temporal window radius $r$ is defined as a distinct scale number that decreases gradually.
We denote the $j$-th layer as $L_j$ and the total number of layers as $M$, and the radius sizes in our cross-modal alignment module can be represented as $R={r_j}_{j=1}^{M}$.
Given the output visual and text embeddings of the $j$-th layer $\boldsymbol{V}^j\in \mathbb{R}^{T \times D}$ and $\boldsymbol{Q}^j\in \mathbb{R}^{L \times D}$, the $(j+1)$-th layer takes them as input to iteratively learn the joint semantic alignment.
The $\boldsymbol{V}^j$ and $\boldsymbol{Q}^j$ are also fed into the boundary detection head to predict the timestamps of anchors with the corresponding scale of the $j$-th layer.
In addition, we evenly allocate $H$ scales to $M$ layers in order of prediction size, and the temporal window radius of each layer is set as its maximum assigned scale.

Unlike most multi-scale mechanism that progressively expands their visual scope to learn low-level semantics in early layers and high-level semantics in later layers, our multi-scale hierarchies work in reverse by \textit{reducing the visual scope from bottom to top}.
This design allows our model to encode the semantics of events with high SNR at an early stage, and  gradually refine the event understanding with low SNR to reduce visual redundancy.

\subsection{Zoom-in Boundary Detection (ZBD)}
Building upon the cross-modal alignment module with MNA, we further propose a two-stage anchor-based detection manner named zoom-in boundary detection (ZBD).

In the first stage, we adopt multi-scale detection to take full advantage of multi-scale representations enhanced by our MNA operator.
Specifically, we first adopt an integration block with single transformer layer to integrate video and text embeddings $\boldsymbol{V}^j$ and $\boldsymbol{Q}^j$.
The integration block takes the visual embedding $\boldsymbol{V}^j$ as query and text embedding $\boldsymbol{Q}^j$ as key/value to generate new visual embedding $\tilde{\boldsymbol{V}^j}$. We then concatenate $\tilde{\boldsymbol{V}^j}$ with $\boldsymbol{V}^j$, resulting in a new embedding  $\hat{\boldsymbol{V}}^j\in \mathbb{R}^{T \times 2D}$ for the $j$-th layer.
Then we use two small MLPs to predict the coordinate offsets and scores of the anchors in $j$-th layer.
Note that the detection heads of different layers share the same model design and parameters, including the integration block, as well as the classification and regression layers.

In the second stage, we start by selecting the top $N$ proposals with the highest scores and extracting their corresponding temporal \textbf{R}egion \textbf{O}f \textbf{I}nternet (ROI) representations.
Next, we sample 3 frame-level embeddings from  $\hat{\boldsymbol{V}}=\{\hat{\boldsymbol{V}}^j\}_{j=1}^{M}$ in propotion to the selected proposals.
We then concatenate the sampled features to represent each selected proposal, denoted as $\boldsymbol{A}=\{A_i\}_{i=1}^{N}\in \mathbb{R}^{N \times 6D}$. 
Based on $\boldsymbol{A}$, we use two additional MLPs to predict offsets and scores of the $N$ proposals with fine-grained adjustment.

\subsection{Training Techniques}
In the first stage, we assign a discrete binary class label to each anchor. 
An anchor ($\hat{t}^s$, $\hat{t}^e$) is considered positive only if its Intersection over Union (IoU) score with the ground truth timestamp ($t^s$, $t^e$) is higher than a certain threshold. 
Following this, we first compute the classification loss as:

\begin{equation}
    \mathcal{L}_{cls}=-\frac{1}{HT}\sum_{i=1}^{HT} t_i\log(p_i) +(1-t_i)\log(1-p_i)
\end{equation}
where $t_i$ is the IoU score label, $p_i$ is the prediction score of the $i$-th anchor.
To enhance our model's grounding ability of start/end timestamps, we further use a temporal boundary loss (smooth $L_1$) and compute the regression loss as:

\begin{equation}
    \mathcal{L}_{reg} = \frac{1}{N_{reg}}\sum_{i=1}^{{N_{reg}}} \mathcal{L}_{L1}(\hat{t}^s_i, t^s_i) + \mathcal{L}_{L1}(\hat{t}^e_i, t^e_i)
\end{equation}

Notice that the regression loss is only activated for positive proposals. 
$N_{reg}$ denotes the number of positive proposals. 
To control the ratio between classification loss and regression loss, we employ a trade-off parameter $\mu$. 
The final objective of the first stage is denoted as: 

\begin{equation}
    \mathcal{L}^1 = \mathcal{L}_{cls}^{1} + \mu \mathcal{L}_{reg}^{1}
\end{equation}

In the second stage, we use the same loss for the selected proposals. 
To speed up convergence, we randomly select $N_{pos}$ positive proposals from the ground truth and add them to the top-$N$ proposal list during the training phase. 
Our training process is end-to-end, and the objective is:

\begin{equation}
    \mathcal{L} = \mathcal{L}^{1} + \lambda\mathcal{L}^{2}
\end{equation}

%% file: section/5-experiment.tex
\input{tabs/comp_sota}
\subsection{Experimental Settings}

\noindent\textbf{Datasets.}
To evaluate our proposed model, we carry out experiments on three widely-used TVG benchmarks. \textbf{(1) ActivityNet}~\cite{Krishna_2017_ANET} contains 20K untrimmed videos with an average of 3.65 clip-query pairs per video. 
Following previous works~\cite{liu2021IANET,zhang20192DTAN,liu2020CSMGAN}, we split the dataset into three subsets: 37,421 queries for training, 17,505 for validation and 17,031 for testing.
\textbf{(2) Charades-STA}~\cite{sigurdsson2016Charades} comprises 6,672 videos depicting daily indoor activities, with 12,408 clip-query pairs for training and 3,720 pairs for testing.
\textbf{(3) Ego4d-NLQ}~\cite{grauman2022ego4d} consists of 1,659 untrimmed videos with an average duration of 500 seconds. 
Each video contains an average of 12 clip-query pairs. 
Following the official split provided by \cite{grauman2022ego4d}, we use 11,291 queries for training, 3,874 for validation and 4,005 for testing.

\noindent \textbf{Implementation Details.}
To ensure a fair comparison, we use the same C3D and I3D features provided by \cite{liu2021IANET,zhang20192DTAN} for ActivityNet and Charades-STA, while for Ego4D-NLQ, we use slowfast features \cite{feichtenhofer2019slowfast} provided by \cite{grauman2022ego4d}. 
We set the max sampling frame number to 200, 64, and 600 for ActivityNet, Charades-STA, and Ego4d-NLQ, respectively. 
Moreover, We use Glove \cite{pennington2014GLOVE} to provide text embeddings for the queries.
Our model adopts light-weight video and text encoders, each with 2 layers.
The cross-modal encoder comprises 4 transformer layers with a hidden size of 512 and a head number of 4. 
Following ~\cite{liu2021IANET}, we set the anchor radius as [4, 8, 16, 32, 48, 64, 80, 96] for ActivityNet, [8, 12, 16, 20] for Charades-STA, and [4, 8] for Ego4D-NLQ. 
Additionally, we set the radius of temporal windows in the cross-modal alignment module as [96, 64, 32, 8], [20, 16, 12, 8], and [32, 16, 8, 4] respectively for the three datasets.
We train our model from scratch using Adam~\cite{Adam} as the optimizer. 
For the hyper-parameters, we set $\lambda$ to 0.1.
For ActivityNet and Ego4D-NLQ, we set $\mu$ to 1e-3, and for Charades, we set it to 5e-3.
We also set $N$ and $N_{pos}$ in the second detection stage to 64 and 4.

\noindent \textbf{Evaluation Metrics.}
Following \cite{liu2021IANET,zhang20192DTAN}, we adopt ``R@n, IoU@m'' as the evaluation metric.
The metric calculates the percentage of at least one of the top-$n$ predicted proposals having IoU greater than $m$ compared to the ground truth timestamps. 
For ActivityNet and Charades-STA, we use $n \in \{1,5\}$ and $m \in \{0.5, 0.7\}$. For Ego4D-NLQ, we use $n \in \{1,5\}$ and $m \in \{0.3, 0.5\}$.

\subsection{Comparison Results}

\noindent\textbf{Comparison of performance.} 
Table~\ref{tab:compare} and~\ref{tab:compare2} present a comparison of the performance of different models on the three datasets, which demonstrates the competitiveness of our model. 
Specifically, on the traditional TVG benchmarks: ActivityNET and Charades-STA, our model significantly surpasses previous works except for SSRN~\cite{zhu2023SSRN} which uses a rule-based approach to generate soft-labels.
When removing the soft-label setting in SSRN, the fair comparison results between our model and SSRN are 49.72/\textbf{51.26} (Ours/SSRN) on R1@0.5, and \textbf{31.25}/31.02 (Ours/SSRN) on R1@0.7. In addition to the comparable SOTA performance, our model has a parameter advantage over SSRN as shown in Table~\ref{tab:compare3}.
Furthermore, our model performs exceptionally well on the challenging Ego4d-NLQ benchmark as shown in Table~\ref{tab:compare2}, achieving $\textbf{+12\%}$ and $\textbf{+30\%}$ relative improvement over previous SOTA on Rank1@0.3 and Rank1@0.5, respectively.

These results demonstrate the robustness of our model on a broad range of benchmarks, particularly on hard cases like Ego4d-NLQ, which has extremely low SNR.

\input{tabs/comp_efficiency}

\noindent\textbf{Comparison of inference efficiency.} 
We follow previous work~\cite{liu2021IANET} and take 200 frame visual features and 20 word embeddings as inputs. 
To evaluate our model efficiency, we compare its running time and model size with those of other models during the inference phase.
All experiments are conducted on a single Nvidia TITAN-XP GPU, with the inference phase being repeated 10,000 times to report the average cost.
Thanks to our light-weight network architecture, we achieve superior performance with relatively few model parameters as shown in Table~\ref{tab:compare3}.

\subsection{Ablation Study}
\noindent\textbf{SNR Analysis. }
Our observation reveals that previous SOTAs cannot address the visual noise issue caused by increasing the sampling rate on Ego4D with extremely low SNR, resulting in performance degradation.
In contrast, our method achieves the best performance even at a sampling rate of 600 frames, where the visual noise is at its maximum.
We further analyze how our proposed components reverse the tendency. 
As shown in Table~\ref{tab:ego4d_2}, the \textbf{NA} (neighboring attention) component is indispensable in filtering out  discriminative information relevant to the query from overwhelming visual noises.
\input{tabs/insight_Ego4d}

\noindent\textbf{Impact of our proposed modules.}
We use the degraded model, which removes all of our proposed modules, as the baseline. We then progressively add the modules to validate their impact in Table~\ref{tab:abl_total}.
Our model achieves significant improvements on all metrics by implementing the neighboring attention (NA) in transformer blocks.
On Charades-STA, it achieves improvements of $\textbf{+1.94}$ Rank1@0.5 and $\textbf{+1.72}$ Rank1@0.7.
The improvement is even more significant  on ActivityNet which has a lower SNR. 
This suggests that our neighboring attention can effectively mitigate the impact of low SNR.
In addition, Row 3 reveals that incorporating a multi-scale feature hierarchy leads to notable performance gains and enhanced visual representation.
We also observe that this multi-scale hierarchy benefits the detection stage, as illustrated in Row 4.
By combining all proposed modules, our model achieves the best performance on both ActivityNet and Charades-STA.

\input{tabs/abl_modules}

\input{tabs/abl_neighbor_attn}

\noindent\textbf{Impact of different MNA settings. }
In Table~\ref{tab:abl_loc2}, we focus on the ablation of two factors in the MNA setting: the temporal window type and radius.
The temporal window includes three distinct types:
\textbf{(1) \emph{Fixed}}, where each layer has a fixed size of temporal window;
\textbf{(2) \emph{Increase}}, where the window size gradually increases; \textbf{(3) \emph{Decrease}}, where the window size gradually decreases.

It can be observed that the best window size is 8 when it is fixed.
Using a multi-scale pyramid with \emph{Decrease} window sizes leads to significantly improved results.
In addition, our approach keeps the sequence length unchanged, both  \emph{Increase} and \emph{Decrease} are able to perform multi-scale feature encoding.
Such approach outperforms single-scale features with a \emph{Fix} window size on detection tasks.

We also compare our MNA module with deformable attention (DFA)~\cite{DeDETR} implemented by \cite{Wang_2021_ICCV_pdvc}.
To ensure a fair comparison, we use the same number of sampling points in the deformable attention as in the MNA. 
We observe that the performance of our model using MNA outperforms the model using DFA, achieving $\textbf{+1.71}$ improvement.
This indicates that our proposed MNA is more effective to aggregate local-wise key information for the TVG task.

\input{tabs/abl_zoom_in}
\noindent\textbf{Impact of different zoom-in detection settings.}
As shown in Table~\ref{tab:abl_twostage}, there are two factors that  significantly impact zoom-in detection: the number of positive samples $N_{pos}$ selected during training and the number of candidates $N$ selected for the zoom-in process.
Our model performs best when $N$=$64$ and $N_{pos}$=$4$.
Note that incorporating positive proposals during training consistently leads to performance improvements. 
We believe this is because the positive proposals help speed up the model training convergence.

\subsection{Further Discussion}
\input{tabs/abl_plug_in}
\noindent\textbf{Is our proposed MNA easy to plug in? }
We conduct experiments on IANet~\cite{liu2021IANET} and Moment-Detr~\cite{lei2021MomentDTER}, both with and without our proposed MNA.
For IANet, we replace its alignment module with our cross-modal alignment module while keeping all other settings the same.
As shown in Table~\ref{tab:abl_loc1}, IANet with our MNA performs significantly better, with Rank1@0.5 improved from 46.69 to 47.43 and Rank1@0.7 from 25.50 to 29.07. 
For Moment-Detr, we simply plug in our MNA module to the transformer blocks without any additional modifications, and the improvement is also promising.
These results validate the portability of our proposed multi-scale neighboring attention.

\begin{figure}[t!]
	\begin{center}
		\includegraphics[width=0.9\linewidth]{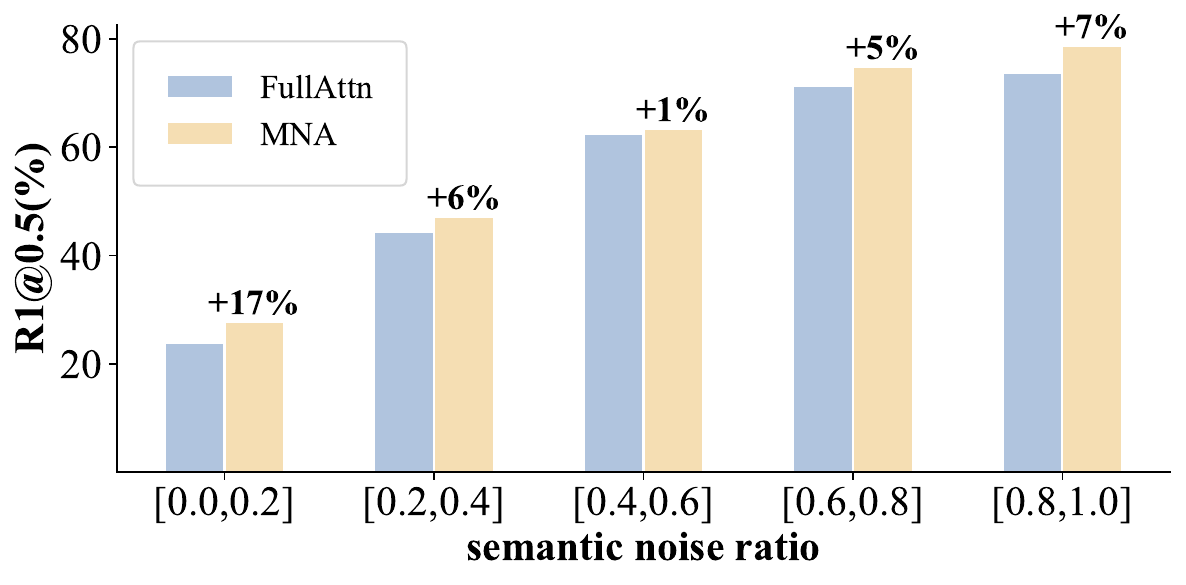}
	\end{center}
     \vspace{-6pt}
	\caption{Ablation of queries across different SNRs.}

	\label{fig:abl_loc}
\end{figure}

\noindent\textbf{What queries can be improved by MNA? }
To answer this question, as shown in Figure~\ref{fig:abl_loc}, we analyze the accuracy distribution of queries across different SNRs on ActivityNet.
The results indicate that queries on videos with the lowest SNR (e.g., [0.0, 0.2]) achieve the largest improvement: $\textbf{+17\%}$, highlighting the efficacy of our MNA approach in mitigating the negative effects of low SNR.

\input{tabs/abl_multiscale}

\noindent\textbf{Can multi-scale hierarchy in encoder benefit the detection stage?} 
Our work utilizes a multi-scale detection scheme in our zoom-in boundary detection module. 
This scheme leverages the extracted information with multi-scale feature hierarchies in the cross-modal alignment module.
We aim to investigate if this scheme can be applied to other cross-modal alignment modules as well.
To answer this question, we perform experiments to validate different combinations of cross-modal alignment modules and the multi-scale detection scheme.
We compare three types of cross-modal alignment modules: (1) IAnet alignment module; (2) a normal transformer cross-encoder module (full attention); (3) our cross-modal alignment module (MNA).
As shown in Table~\ref{tab:abl_loc3},  the combination of the multi-scale  detection scheme and our alignment module with MNA achieves the greatest improvement. 
It indicates that our encoding methodology with proposed multi-scale hierarchies can bring further enhancement in the detection phase, resulting in superior performance.

\noindent\textbf{Differences from previous local attention methods?} Instead of previous local attention which focuses on a single modality (e.g., Longformer~\cite{beltagy2020longformer} and BigBird~\cite{zaheer2020big} address the problem of long-distance dependency within a single modality with a trade-off between performance and efficiency), our MNA takes full advantage of the characteristics of TVG task (the semantics of the text query only corresponds to a small part of the whole video) to enhance the semantic alignment of two asymmetric modalities, and achieves a win-win situation of performance and efficiency. 

%% file: tabs/comp_sota.tex
\section{Experiments}
\begin{table*}[ht]
    \small
    \begin{center}
    \setlength{\tabcolsep}{1mm}{
    \begin{tabular}{c|c|ccccc|ccccc}
    \hline \hline
    \multirow{2}*{Methods}&\multirow{2}*{Venue} & \multicolumn{5}{c|}{ActivityNet} & \multicolumn{5}{c}{Charades-STA} \\ \cline{3-12}
    ~&~ & Feats.&R1@0.5 & R1@0.7 & R5@0.5 & R5@0.7 & Feats. &R1@0.5& R1@0.7 & R5@0.5 & R5@0.7 \\ 
    \hline
    CTRL~\cite{gao2017CTRL}&ICCV2017 &C3D & 29.01 & 10.34 & 59.17 & 37.54 & C3D&23.63 & 8.89 & 58.92 & 29.57 \\
    ACRN~\cite{liu2018ACRN} &SIGIR2018&C3D & 31.67 & 11.25 & 60.34 & 38.57  &C3D & 20.26 & 7.64 & 71.99 & 27.79 \\
    SCDM~\cite{yuan2019SCDM} &NeurIPS2019&C3D & 36.75 & 19.86 & 64.99 & 41.53& I3D&54.44 & 33.43 & 74.43 & 58.08 \\
    LGI~\cite{mun2020LGI}&CVPR2020 &C3D & 41.51 & 23.07 & - &  - & I3D & 59.46 & 35.48 & - & - \\
    DRN~\cite{zeng2020DRN} &CVPR2020&C3D & 45.45 & 24.36 & 77.97 & 50.30 & I3D &53.09 & 31.75 & 89.06 & 60.05 \\
    VLSNet~\cite{zhang2020VLSNET}&ACL2020 &C3D & 43.22 & 26.16 & - & -  & C3D & 47.31 & 30.19 & - & - \\
    2DTAN~\cite{zhang20192DTAN}&AAAI2020 &C3D & 44.51 & 26.54 & 77.13 & 61.96 & - & - & - & - & - \\
    CSMGAN~\cite{liu2020CSMGAN} &ACMMM2020&C3D & 49.11&29.15&77.64&59.63&I3D&60.04&37.34&89.01&61.85\\
    MS2DTAN~\cite{zhang2021ms2dtan} &TPAMI2021&C3D & 46.16	& 29.21 & 78.80 & 60.85 & C3D & 41.10 & 23.25 & 81.53 & 48.55 \\
    BPNet~\cite{xiao2021BPNET}&AAAI2021 &C3D & 42.07 & 24.69 & - & - & I3D & 50.75 & 31.64 & - & - \\
    CBLN~\cite{liu2021CBLN}&CVPR2021 &C3D & 48.12 & 27.60 & 79.32 & 63.41 & I3D &61.13 & 38.22 & 90.33 & 61.69 \\
 
    MSA~\cite{zhang2021MSA} &CVPR2021&C3D & 48.02&\underline{31.78}&78.02&63.18&-&-&-&-&-\\

    IANet~\cite{liu2021IANET} &EMNLP2021&C3D & 48.57 & 27.95 & 78.99& 63.12& I3D&61.29& 37.91 &89.78 &62.04 \\
    TBP~\cite{hao2022TBP}&ECCV2022&C3D&43.11&25.84&-&-&I3D&57.69&39.27&-&-\\
    MMN~\cite{wang2022MMN} &AAAI2022 &C3D & 48.59&29.26&79.50&64.76&-&-&-&-&-\\
    SSRN~\cite{zhu2023SSRN}&EMNLP2022 &C3D & \textbf{54.49} & \textbf{33.15} &\textbf{84.72} & \textbf{68.48}& I3D&\textbf{65.59}& \textbf{42.65} &\textbf{94.76} &\textbf{65.48} \\
    
    \hline
    \multicolumn{2}{c|}{\multirow{2}*{\textbf{ours}}}&C3D & \underline{49.72} & 31.25&\underline{80.13}&\underline{65.56}&C3D&49.78&31.05&86.77&56.02 \\
    \multicolumn{2}{c|}{~} &I3D & 51.38 & 32.10&83.23&67.04&I3D&\underline{65.21}&\underline{40.59}&\underline{92.22}&\underline{63.15} \\
    \hline \hline
    \end{tabular}
    }
    \end{center}
        \vspace{-6pt}
    \caption{Performance comparison on ActivityNet and Charades-STA.}
    \vspace{-5pt}
    \label{tab:compare}
\end{table*}

\begin{table}[t!]
    \small
    \begin{center}
    \setlength{\tabcolsep}{1.5mm}{
    \begin{tabular}{c|cccc}
    \hline \hline
    \multirow{3}*{Method} & \multicolumn{4}{c}{ Ego4D}  \\ \cline{2-5}
    ~ & R@1, & R@5, & R@1, & R@5, \\ 
    ~ & IoU=0.3 & IoU=0.3 & IoU=0.5 & IoU=0.5 \\ \hline
   Moment-Detr~\cite{lei2021MomentDTER}\dag & 4.33 & 6.12 & 1.21 & 3.78 \\
   MSA~\cite{zhang2021MSA}\dag & 3.09 &8.11 & 1.41 & 4.05 \\
   IANet~\cite{liu2021IANET}\dag& 4.77& 14.01 & 2.54 & 7.12 \\
   MMN~\cite{wang2022MMN}\dag& 6.03& 13.53 & 3.41 & 6.57 \\
   2DTAN~\cite{zhang20192DTAN}$\ast$ & 5.04 & 12.89 & 2.02 & 5.88 \\
    VLSNet~\cite{zhang2020VLSNET}$\ast$ & 5.45& 10.74 &3.12 & 6.63 \\
    \hline
    \rowcolor{Gray}
    \textbf{ours} & \textbf{6.74} & \textbf{16.89} & \textbf{4.06} & \textbf{10.43} \\ 
    \hline \hline
    \end{tabular}}
    \end{center}
        \vspace{-6pt}
    \caption{Performance comparison on Ego4D-NLQ validation set, where $\ast$ denotes the results are reported by \cite{grauman2022ego4d} and \dag~denotes the results are re-implemented by their officially released code.}
    \label{tab:compare2}
    \vspace{-8pt}
\end{table}

%% file: tabs/comp_efficiency.tex
\begin{table}[t!]
    \begin{center}
    \setlength{\tabcolsep}{1.5mm}{
    \begin{tabular}{c|c|c|c}
    \hline \hline
    Method & Runtime & Model Size & Rank1@0.7 \\ \hline
    CTRL~\cite{gao2017CTRL} & 2.23s & 22M &10.34 \\ 
    ACRN~\cite{liu2018ACRN} & 4.31s & 128M &11.25 \\
    2DTAN~\cite{zhang20192DTAN} & 0.57s & 232M &26.54\\ 
    IANet~\cite{liu2021IANET} & 0.11s & 68M &27.95\\ 
    TBP~\cite{hao2022TBP} &0.05s&\textbf{14M}&25.84\\
    MMN~\cite{wang2022MMN} & 0.83s & 152M &29.26\\ 
    SSRN~\cite{zhu2023SSRN} & -& 184M &\textbf{33.15}\\ 
    \hline
    \rowcolor{Gray}
    \textbf{ours} & \textbf{0.05s} & \underline{16M}& \underline{31.25} \\
    \hline \hline
    \end{tabular}}
    \end{center}
     \vspace{-5pt}
     \caption{Comparison of inference efficiency on ActivityNet.}
    \label{tab:compare3}
    \vspace{-5pt}
\end{table}

%% file: tabs/insight_Ego4d.tex
\begin{table}[t!]
    \small
    \begin{center}
    \setlength{\tabcolsep}{4mm}{
    \begin{tabular}{l|ccc}
    \hline \hline
    \multirow{2}*{Method} &  \multicolumn{3}{c}{Number of sampled frames} \\ \cline{2-4}
    ~ & 200 & 400 & 600\\ \hline
   Moment-Detr~\cite{lei2021MomentDTER}\dag & 4.33&3.24&1.97\\
   MSA~\cite{zhang2021MSA}\dag &3.09 & 2.54 & 2.35\\
   IANet~\cite{liu2021IANET}\dag& 4.77& 4.01 & 3.21 \\
   VLSNet~\cite{zhang2020VLSNET}$\ast$ & 5.45& 5.13 &4.32 \\
    \hline
    ours & \textbf{5.71}&\textbf{6.02}&\textbf{6.74}\\ 
    ours - NA & 4.47&3.52&2.44\\ 
    ours - MS & 5.55&6.07&6.65\\ 
    ours - Zoom-in & 5.08&5.89&6.42\\ 
    \hline \hline
    \end{tabular}}
    \end{center}
        \vspace{-3pt}
    \caption{Rank1@0.3 accuracy on Ego4D-NLQ validation set with different number of sampled frames, where ``-" denotes removing the corresponding component from our model, ``NA" denotes neighboring attention, ``MS" denotes multi-scale.}
    \label{tab:ego4d_2}
    \vspace{-6pt}
\end{table}

%% file: tabs/abl_modules.tex
\begin{table}[t!]
    \small
    \begin{center}
    \setlength{\tabcolsep}{1.5mm}{
    \begin{tabular}{l|l|cc|cc}
    \hline \hline
   & \multirow{2}*{Method} & \multicolumn{2}{c|}{ActivityNet} & \multicolumn{2}{c}{Charades-STA} \\ \cline{3-6}
   & ~ & R1@0.5 & R1@0.7 & R1@0.5 & R1@0.7\\ 
    \hline
   1& Baseline &44.62&23.70&58.23&34.03 \\
    
   2& + NA &46.68&28.64&60.17&35.75 \\
   3& + MS enc &47.75&29.11&60.61&37.06\\
   4& + MS det &48.97&30.36&62.48&37.86\\
    \rowcolor{Gray}
   5& + Zoom-in & 49.72 & 31.25  & 65.21 & 40.59 \\
    \hline \hline
    \end{tabular}}
    \end{center}
        \vspace{-6pt}
    \caption{Ablation of different proposed modules. ``NA" denotes neighboring attention. ``MS" denotes multi-scale}
    \label{tab:abl_total}
\end{table}

%% file: tabs/abl_neighbor_attn.tex
\begin{table}[t!]
    \small
    \begin{center}
    \setlength{\tabcolsep}{2mm}{
    \begin{tabular}{c|c|c|c}
    \hline \hline
    \multirow{2}*{Method} & \multirow{2}*{Type Size} & \multirow{2}*{Radius}& \multicolumn{1}{c}{ActivityNet} \\\cline{4-4}
      & & & R1@0.5\\ 
    \hline
    MNA & \emph{Fixed}&[4, 4, 4, 4]&45.65\\
    \rowcolor{Gray}
    MNA & \emph{Fixed}&[8, 8, 8, 8]&\textbf{46.68} \\
    MNA & \emph{Fixed} &[32, 32, 32, 32]&46.03 \\
    MNA & \emph{Fixed} &[64, 64, 64, 64]&45.16 \\
    MNA & \emph{Fixed}&[96, 96, 96, 96]&45.00 \\
    \hline
    MNA & \emph{Fixed}&[8, 8, 8, 8] &46.68\\
    MNA & \emph{Increase}&[8, 32, 64, 96] & 47.03 \\
    \rowcolor{Gray}
    MNA & \emph{Decrease}&[96, 64, 32, 8] &\textbf{47.75} \\
    \hline
    DFA &  \emph{Fixed} &[8, 8, 8, 8]&44.97 \\
    \rowcolor{Gray}
    MNA & \emph{Fixed}&[8, 8, 8, 8] &\textbf{46.68}\\

    \hline \hline
    \end{tabular}}
    \end{center}
        \vspace{-6pt}
    \caption{Ablation of different neighboring attention settings. MNA denotes our proposed neighboring attention. DFA denotes  the deformable attention. Note that the multi-scale detection and zoom-in process are not employed in this experiment.}
    \label{tab:abl_loc2}
    \vspace{-6pt}
\end{table}

%% file: tabs/abl_zoom_in.tex
\begin{table}[t!]
    \small
    \begin{center}
    \setlength{\tabcolsep}{3mm}{
    \begin{tabular}{c|c|c|cc}
    \hline \hline
    \multirow{2}*{Row} & \multirow{2}*{top-N} & \multirow{2}*{$N_{pos}$}& \multicolumn{2}{c}{ActivityNet} \\\cline{4-5}
      & & & R1@0.5& R1@0.7\\ 
    \hline
    1 & 32&0&48.64 & 29.90\\
    \rowcolor{Gray}
    2 & 64&0&49.13 & 30.44 \\
    3 & 128 &0&48.34&29.98 \\
    4 & 256 &0&48.56&30.21 \\
    \hline
    5 & 64& 0 &49.13 & 30.44\\
    \rowcolor{Gray}
    6 & 64&4 &49.72& 31.25 \\
    7 & 64&8 &49.52& 30.67 \\
    8 & 64&16 &49.40&30.78 \\
    
    \hline \hline
    \end{tabular}}
    \end{center}
        \vspace{-6pt}
    \caption{Ablation of different zoom-in detection settings, where top-N and $N_{pos}$ denote the number of selected candidates and selected positive samples for zoom-in process and training stage.}
    \label{tab:abl_twostage}
\end{table}

%% file: tabs/abl_plug_in.tex
\begin{table}[t!]
    \small
    \begin{center}
    \setlength{\tabcolsep}{1.5mm}{
    \begin{tabular}{c|c|ll}
    \hline \hline
    \multirow{2}*{Method} & \multirow{2}*{NeAttn} & \multicolumn{2}{c}{ActivityNet} \\ \cline{3-4}
      & & R1@0.5 & R1@0.7\\ 
    \hline
    \multirow{2}*{IANet~\cite{liu2021IANET}} &\ding{55}& 46.69&25.50\\
    
    ~ &  \ding{51}& $47.43 (\textbf{\color{blue}{+0.7}}) $&$29.07  (\textbf{\color{blue}{+3.6}}) $\\
    \hline
     \multirow{2}*{Moment-Detr~\cite{lei2021MomentDTER}} &\ding{55}& 43.41&23.86\\
    ~ & \ding{51}& $45.13 (\textbf{\color{blue}{+1.7}}) $&$25.34 (\textbf{\color{blue}{+1.5}}) $\\
    
    \hline \hline
    \end{tabular}}
    \end{center}
        \vspace{-6pt}
    \caption{Ablation to demonstrate that our proposed neighboring attention mechanism is easy to plug in other works.}
    \label{tab:abl_loc1}
    \vspace{-3mm}
\end{table}

%% file: tabs/abl_multiscale.tex
\begin{table}[t!]
    \small
    \begin{center}
    \setlength{\tabcolsep}{2mm}{
    \begin{tabular}{c|c|ll}
    \hline \hline
    \multirow{2}*{Method} & \multirow{2}*{multi-scale in det} & \multicolumn{2}{c}{ActivityNet} \\ \cline{3-4}
      & & R1@0.5 & R1@0.7\\ 
    \hline
    \multirow{2}*{IANet}&\ding{55}& 46.69&25.50 \\
     ~&\ding{51}& $45.04 (\textbf{\color{red}{-1.6}}) $&$24.72 (\textbf{\color{red}{-0.8}}) $ \\
    \hline
    \multirow{2}*{FullAttn}&\ding{55}& 44.62&23.70  \\
    ~&\ding{51}& $44.81 (\textbf{\color{blue}{+0.2}}) $& $24.06 (\textbf{\color{blue}{+0.4}}) $ \\
      \hline
    \multirow{2}*{MNA}&\ding{55}& 47.75&29.11 \\
    ~&\ding{51}& $48.97 (\textbf{\color{blue}{+1.2}}) $&$30.36 (\textbf{\color{blue}{+1.3}}) $ \\

    \hline \hline
    \end{tabular}}
    \end{center}
    \vspace{-6pt}
    \caption{Ablation of different combinations of alignment module and multi-scale detection scheme.}
    \label{tab:abl_loc3}
    \vspace{-5pt}
\end{table}

%% file: section/6-conclusion.tex
\section{Conclusion}
In this work, we propose a simple yet effective model for Temporal Video Grounding. 
To mimic human cognitive habits, we first propose a multi-scale neighboring attention mechanism to effectively align the untrimmed video and language query by controlling the visual aggregation from neighboring contexts. 
Additionally, we propose a zoom-in boundary detection module to screen out the top candidates and focus on local-wise discrimination, allowing for fine-grained grounding adjustment.
Our approach achieves competitive performance with around 2X inference speedup and 75\% parameter savings.
